# Iterative Augmentation of Visual Evidence for Weakly-Supervised Lesion Localization in Deep Interpretability Frameworks: Application to Color Fundus Images

Cristina González-Gonzalo, Bart Liefers, Bram van Ginneken, and Clara I. Sánchez

***Abstract*— Interpretability of deep learning (DL) systems is gaining attention in medical imaging to increase experts' trust in the obtained predictions and facilitate their integration in clinical settings. We propose a deep visualization method to generate interpretability of DL classification tasks in medical imaging by means of visual evidence augmentation. The proposed method iteratively unveils abnormalities based on the prediction of a classifier trained only with image-level labels. For each image, initial visual evidence of the prediction is extracted with a given visual attribution technique. This provides localization of abnormalities that are then removed through selective inpainting. We iteratively apply this procedure until the system considers the image as normal. This yields augmented visual evidence, including less discriminative lesions which were not detected at first but should be considered for final diagnosis. We apply the method to grading of two retinal diseases in color fundus images: diabetic retinopathy (DR) and age-related macular degeneration (AMD). We evaluate the generated visual evidence and the performance of weakly-supervised localization of different types of DR and AMD abnormalities, both qualitatively and quantitatively. We show that the augmented visual evidence of the predictions highlights the biomarkers considered by experts for diagnosis and improves the final localization performance. It results in a relative increase of 11.2±2.0% per image regarding sensitivity averaged at 10 false positives/image on average, when applied to different classification tasks, visual attribution techniques and network architectures. This makes the proposed method a useful tool for exhaustive visual support of DL classifiers in medical imaging.**

Manuscript received April 17, 2020; accepted May 10, 2020. Date of publication May 28, 2020; date of current version October 28, 2020. This work was supported by the Dutch Technology Foundation STW, which is part of the Netherlands Organisation for Scientific Research (NWO) and partly funded by the Ministry of Economic Affairs (Perspectief programme P15-26 'DLMedIA: Deep Learning for Medical Image Analysis'). *(Corresponding author: Cristina González-Gonzalo.)*

Cristina González-Gonzalo and Bart Liefers are with the A-Eye Research Group, Diagnostic Image Analysis Group, Department of Radiology and Nuclear Medicine, Radboudumc, 6525 Nijmegen, The Netherlands, and also with the Donders Institute for Brain, Cognition and Behavior, Radboudumc, 6525 Nijmegen, The Netherlands (e-mail: cristina.gonzalezgonzalo@radboudumc.nl; bart.liefers@radboudumc.nl).

Bram van Ginneken is with the Diagnostic Image Analysis Group, Department of Radiology and Nuclear Medicine, Radboudumc, 6525 Nijmegen, The Netherlands (e-mail: bram.vanginneken@radboudumc.nl).

Clara I. Sánchez is with the A-Eye Research Group, Diagnostic Image Analysis Group, Department of Radiology and Nuclear Medicine, Radboudumc, 6525 Nijmegen, The Netherlands, also with the Donders Institute for Brain, Cognition and Behaviour, Radboudumc, 6525 Nijmegen, The Netherlands, and also with the Department of Ophthalmology. Radboudumc, 6525 Nijmegen, The Netherlands (e-mail: clara.sanchezgutierrez@radboudumc.nl).

## I. Introduction

DEEP learning (DL) systems in medical imaging have shown to provide high-performing approaches for diverse classification tasks in healthcare, such as screening of eye diseases [1], [2], scoring of prostate cancer [3], or detection of skin cancer [4]. Nevertheless, DL systems are often referred to as "black boxes" due to the lack of interpretability of their predictions. This is problematic in healthcare applications [5], [6], and hinders experts' trust and the integration of these systems in clinical settings as support for grading, diagnosis and treatment decisions. There is thus an increasing demand for interpretable systems in medical imaging that could further explain models' decisions. Defining an interpretability framework as the combination of a DL system to perform a classification task and a procedure for generating explainable predictions, several such frameworks have been proposed in different medical applications and imaging modalities [4], [7]–[17].

Among the integrated procedures, those based on visual attribution have become very popular, such as the ones defined and described in Table I: saliency [18], guided backpropagation [19], integrated gradients [20], Grad-CAM [21], and guided Grad-CAM [21]. These *attribution methods* provide an interpretation of the network's decision by assigning an attribution value, sometimes also called "relevance" or "contribution", to each input feature of the network depending on its estimated contribution to the network output [22]. This allows to highlight features in the input image that contribute to the output prediction and, consequently, the weakly-supervised detection of objects. However, it has been shown for natural images that these methods localize only the most discriminative parts of an object, instead of highlighting all parts that are relevant for a complete explainability [23], [24].

TABLE I
IMPLEMENTED VISUAL ATTRIBUTION METHODS

| Name | Definition | Description |
|---|---|---|
| Saliency [18] | $M_{SAL} = \frac{\partial \mathcal{F}_{cnn}(I)}{\partial I}$ | It indicates which local morphology changes in the image would lead to modifications in the network's prediction. |
| Guided backpropagation [19] | $M_{GBP} = \frac{\partial \mathcal{F}_{cnn}(I)}{\partial I}$ s.t. $R_i^l = 1_{R_i^{l+1}>0} 1_{f_i^l>0} R_i^{l+1}$, where $R_i^{l+1} = \frac{\partial \mathcal{F}_{cnn}(I)}{\partial f_i^{l+1}}$, $f_i^{l+1} = ReLU(f_i^l)$ and $f_i^l$ is the *i-th* feature map at convolutional layer $l$ | It provides additional guidance to the signal backpropagated through ReLU activations from the higher layers, preventing backward stream of gradients associated to neurons that decrease the activation of the output node. |
| Integrated gradients [20] | $M_{IG} = (I - \bar{I}) \int_0^1 \frac{\partial \mathcal{F}_{cnn}(\bar{I}+\alpha(I-\bar{I}))}{\partial I} d\alpha$ | The generated maps measure the contribution of each pixel in the input image to the prediction. Instead of computing only the gradient with respect to the current input value, this method computes the average gradient while the input varies linearly in several steps from a baseline image (commonly, all zeros) to their current value. |
| Grad-CAM [21] | $M_{G-CAM} = ReLU(\sum_i \alpha_i^l f_i^l)$, where $\alpha_i^l = GAP(\frac{\partial \mathcal{F}_{cnn}(I)}{\partial f_i^l})$, $f_i^l$ is the *i-th* feature map at convolutional layer $l$ and $GAP$ is the global average pooling operation over the two spatial dimensions | The gradients backpropagated from the output to a selected convolutional layer are used for computing a linear combination of the forward activation maps of that layer. Only the pixels with positive influence on the output are maintained, and then rescaled to the input size. |
| Guided Grad-CAM [21] | $M_{GG-CAM} = M_{G-CAM} M_{GBP}$ | It combines guided backpropagation and Grad-CAM, in order to improve the localization ability of the latter method. |

When it comes to medical imaging, the integration of attribution methods allows for the identification of regions discriminant for the final diagnosis. This leads to weakly-supervised localization of abnormalities, which can provide a clinical explanation of the classification output without the need for costly lesion-level annotations. Classification of disease severity in color fundus (CF) images, the focus of this paper, is one medical application where attribution methods have been applied to generate explainable DL predictions and weakly-supervised detection of retinal lesions. In [11] and [12], saliency maps [18] were applied to justify decisions on diabetic retinopathy (DR) and age-related macular degeneration (AMD) classification tasks, respectively. In [13], integrated gradients [20] was used to generate heatmaps for the explanation of predicted DR severity levels. Class activation maps (CAM) [25] were extracted in [14] and [15] also for interpretability of DR diagnosis.

Although these interpretability frameworks have succeeded at localizing abnormal areas related to the predicted diagnosis, the aforementioned limitation of visual attribution methods translates into the clinical domain. Since they localize only the most significant regions, lesions that have less influence on the classification result are ignored, although they could be still important for disease understanding and grading [12], [26]. For some medical imaging modalities and applications, interpretability of abnormal predictions requires the localization of different types of lesions of varying appearance and histologic composition that can be simultaneously present and be responsible for the predicted diagnosis. To overcome this, in [11] and [12] different classifiers are used in parallel, which yields localization of different types of abnormalities in separate maps. This allows for differentiation of abnormalities, but each input image must be processed several times and the interpretability of the actual disease grading remains unclear. Alternatively, to improve lesion localization, some frameworks add customized postprocessing steps [11] or fine-tuning [15] to the attribution methods; or propose tailored architectures with additional interpretation modules [16], [17]. Nevertheless, this conflicts with directly obtaining interpretability of the DL system and hinders the adaptability and generalization among DL classifiers and medical applications.

In this paper, we propose a novel deep visualization method, as an extension to [27], that iteratively unveils abnormalities responsible for anomalous predictions in order to generate a map of augmented visual evidence for DL-based classifiers in medical imaging. This is achieved by combining visual attribution and selective inpainting. The abnormal regions highlighted by visual attribution are inpainted with surrounding local information, guiding the attention to new relevant areas in each iteration. This process also leads to a gradual decrease in the classifier's disease severity prediction, allowing to stop the iterative augmentation once the classification converges to a healthy prediction.

We introduce for the first time the use of selective inpainting for weakly-supervised lesion localization, in order to overcome the main limitation of visual attribution methods [12], [23], [24], [26] and highlight less discriminative areas that might also be relevant for the final diagnosis, locating abnormalities of different types, shapes and sizes. In the proposed method, we integrate an unsupervised technique [28] to perform the selective inpainting.

Defined as a general approach, the proposed iterative method is meant to be seamlessly integrated in diverse interpretability frameworks with different DL classifiers and visual attribution techniques, and without the need of additional customized steps.

We apply the proposed method for the interpretation of automated grading in CF images of two retinal diseases: DR and AMD [29], [30]. For each diagnosis task, we classify images by disease severity and analyze the interpretability performance when the proposed iterative augmentation is applied. We validate the initial and augmented visual evidence maps qualitatively and, in contrast to most previous approaches, we evaluate the performance for weakly-supervised localization of DR and AMD abnormalities quantitatively. We show that the method can be integrated with different visual attribution techniques and different DL classifiers.

Our main contributions can be summarized as follows:

- We propose a novel iterative method for exhaustive explainability of DL-based classification tasks in medical imaging which combines the extraction of visual attribution with selective inpainting, defining a sensible stopping criterion based on the gradual decrease of the predicted disease severity. It is the first time that inpainting is used in weakly-supervised lesion localization.

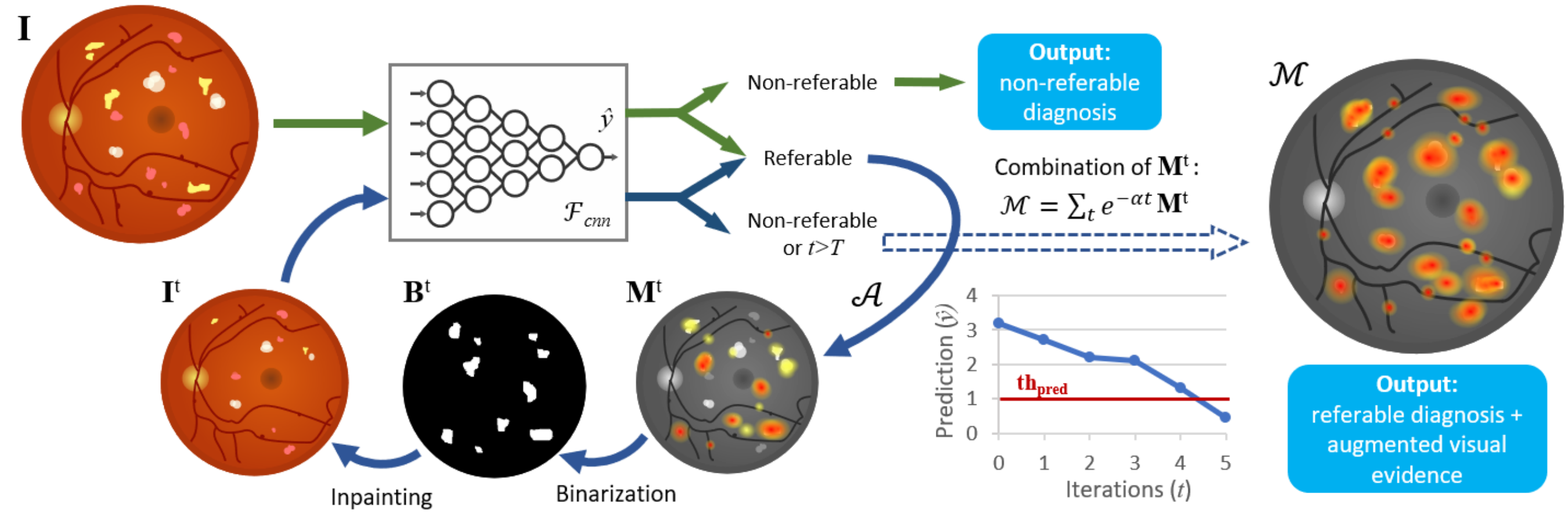

Fig. 1. Overview of the proposed method applied to automated grading of diabetic retinopathy in color fundus images. The workflow to generate the original prediction is depicted in green; the workflow for the proposed iterative visual evidence augmentation is depicted in blue. **I**, input image; $\mathcal{F}_{cnn}$, convolutional neural network; $\hat{y}$, prediction; $th_{pred}$, prediction threshold; $t$, number of iteration; $T$, maximum number of iterations; $\mathcal{A}$, attribution method; $\mathbf{M}^t$, explanation map at iteration $t$; $\mathbf{B}^t$, binary mask at iteration $t$; $\mathbf{I}^t$, inpainted image at iteration $t$; $\mathcal{M}$, augmented explanation map.

- Compared with our conference version [27], we present an improved methodology in terms of generalizability, validated both qualitatively and quantitatively. We present a quantitative comparison of baseline visual attribution methods for weakly-supervised lesion localization and an extensive evaluation of the proposed method, analyzing its agnosticism when applied to different classification tasks (automated grading of DR and AMD in CF images), visual attribution techniques, and network architectures.
- We perform the first quantitative validation of weakly-supervised lesion-level detection for AMD classification.

## II. Methods

The first part of this section describes the proposed iterative visual evidence augmentation, depicted in Fig. 1. The proposed method iteratively unveils areas relevant for a final diagnosis, so as to generate exhaustive visual evidence of classification predictions and, consequently, weakly-supervised lesion-level localization. The second part of the section describes the image-level classification used to provide the DL-based decisions to be interpreted.

### A. Iterative Visual Evidence Augmentation

Let $\mathbf{I} \in \mathbb{R}^{m\times n\times 3}$ be an image with size $m \times n$ pixels (and 3 color channels) and a corresponding label $y$, $\mathcal{F}_{cnn}$ : $\mathbf{I} \longrightarrow \hat{y} \in \mathbb{R}$ a convolutional neural network (CNN) optimized for a classification task using a development set $\mathcal{I} = \{(\mathbf{I}_1, y_1), \ldots, (\mathbf{I}_s, y_s)\}$, and $\mathcal{A} : (\mathbb{R}^{m\times n\times 3}, \mathcal{F}_{cnn}) \longrightarrow \mathbf{M} \in \mathbb{R}^{m\times n}$ an attribution method, such as the ones defined in Table I. For a given $\mathbf{I}$, a prediction $\hat{y}$ is obtained with $\mathcal{F}_{cnn}$. If the image is considered *abnormal* (or referable in the case of retinal images), an explanation map $\mathbf{M}$ is generated by applying $\mathcal{A}$, highlighting areas of $\mathbf{I}$ that are discriminant for $\hat{y}$.

The explanation map $\mathbf{M}$ is binarized to identify the areas where selective inpainting is then applied, in order to remove abnormalities that have been already localized. This procedure is applied iteratively to increase attention to less discriminative areas and generate an augmented explanation map $\mathcal{M}$, by increasing the *normality* of the input image in each iteration. Algorithm 1 includes the pseudocode to calculate the augmented visual evidence, and Fig. 1 shows an overview of the proposed method.

In this work, *normality* is defined based on the predicted value $\hat{y} = \mathcal{F}_{cnn}(\mathbf{I})$, such that an image is considered *normal* (or non-referable in the case of retinal images) if $\hat{y} < th_{pred}$. The prediction threshold $th_{pred}$ is defined in a validation subset of $\mathcal{I}$ by means of Receiver Operating Characteristic (ROC) analysis. The maximum number of iterations $T$ was set to 20. Regarding binarization of the explanation maps, we use the Otsu method [31] to compute $th_{bin}$ and yield an adaptative thresholding. For selective inpainting, we use the Navier-Stokes method [28] with a radius $r_{inp}$ of size 3, based on fluid dynamics to match gradient vectors around the boundaries of the region to be inpainted. The final augmented explanation map $\mathcal{M}$ is obtained by an exponentially decaying weighted sum of the iteratively generated maps $\mathbf{M}$, with $\alpha = 0.6$.

### B. Image-Level Classification

The proposed iterative visual evidence augmentation must be built upon a DL classifier that reaches acceptable performance, so as to achieve reliable interpretability. $\mathcal{F}_{cnn}$ was therefore optimized for each classification task: classification of CF images for detection of DR ($\mathcal{F}^{DR}_{cnn}$) and AMD ($\mathcal{F}^{AMD}_{cnn}$).

Prior to classification, every CF image goes through a preprocessing stage, where the bounding box of the field of view is extracted, then rescaled to 512 × 512 pixels, and lastly, contrast-enhancement based on [32] is applied to reduce local differences in lighting and among images. The contrast-enhanced image is used as input for the classifier.

**Algorithm 1** Iterative Visual Evidence Augmentation

**Input:** Input image $\mathbf{I}$
Trained CNN classifier $\mathcal{F}_{cnn}$
Prediction threshold $th_{pred}$
Maximum number of iterations $T$
Selective inpainting radius $r_{inp}$
**Output:** Augmented explanation map $\mathcal{M}$

Initialize $t = 0$ and $\mathbf{I}^t = \mathbf{I}$
Calculate initial prediction $\hat{y} = \mathcal{F}_{cnn}(\mathbf{I}^t)$

**while** $(\hat{y} \geq th_{pred})$ & $(t < T)$ **do**
$\mathbf{M}^t = \mathcal{A}(\mathbf{I}^t, \mathcal{F}_{cnn})$
Binarize $\mathbf{M}^t$ by thresholding:
$$\mathbf{B}^t(x, y) = \begin{cases} 1, & \text{if } \mathbf{M}^t(x, y) \geq th_{bin}, \\ 0, & \text{otherwise} \end{cases}$$
Inpaint $\mathbf{I}^t$ given mask $\mathbf{B}^t$:
$$\mathbf{I}^t = Selective\ inpainting(\mathbf{I}^t, \mathbf{B}^t, r_{inp})$$
Calculate new prediction $\hat{y} = \mathcal{F}_{cnn}(\mathbf{I}^t)$
$t = t + 1$
**end while**

Compute augmented explanation map $\mathcal{M}$:
$$\mathcal{M} = \sum_t e^{-\alpha t}\mathbf{M}^t$$

The CNNs were based on the VGG-16 architecture [33], pre-trained on ImageNet. They were adapted to input images of size 512 × 512 by applying a stride of 2 in the first layer of the first convolutional block, and using a valid instead of padded convolution for the first layer of the last convolutional block. Dropout layers (p = 0.5) were added in between the fully-connected layers. We followed a regression approach in which the output of a network consists of a single node, representing a continuous value which is monotonically related to predicted disease severity. The loss was defined as the mean squared error between the prediction and the reference-standard label. For each classification task, the optimal classifier $\mathcal{F}_{cnn}$ was selected regarding the performance on a validation set by means of receiver operating characteristic (ROC) analysis, computing the area under the ROC curve (AUC), in order to assure good discrimination between referable and non-referable cases. Additionally, the ability to discriminate between disease stages was measured by means of the quadratic Cohen's weighted kappa coefficient ($\kappa$) [34]. Sensitivity (SE) and specificity (SP) were computed at the optimal operating point of the system, which was considered to be the best tradeoff between the two values, i.e., the point closest to the upper left corner of the graph. This allowed for extraction of the optimal threshold $th_{pred}$ for referability in the corresponding validation set.

## III. Data

### A. Image-Level Classification

The Kaggle DR dataset [35] was used for training, validation and testing of $\mathcal{F}_{cnn}^{DR}$. Images were acquired by different CF digital cameras with varying resolution. Each image was graded by DR severity by a human reader, regarding the International Clinical Diabetic Retinopathy (ICDR) severity scale [36], with stages 0 (no DR), 1 (mild non-proliferative DR), 2 (moderate non-proliferative DR), 3 (severe non-proliferative DR), and 4 (proliferative DR). Categories 0 and 1 are considered non-referable DR and categories 2 to 4 referable DR. This database is divided in two sets: the Kaggle training set (35,126 images from 17,563 patients; one photograph per eye) and the Kaggle test set (53,576 images from 26,788 patients; one photograph per eye).

The classifier for AMD, $\mathcal{F}_{cnn}^{AMD}$, was trained, validated and tested on the Age-Related Eye Disease Study (AREDS) dataset [37]. AREDS was designed as a long-term prospective study of AMD development and cataract in which patients were examined on a regular basis and followed up to 12 years. The AREDS dbGaP set includes digitized CF images. In 2014, over 134,000 macula-centered CF images from 4,613 participants were added to the set (for each patient-visit available, one photograph per eye with their corresponding stereo pairs). We excluded images regarding the criteria in the AREDS dbGaP guidelines [37], and 133,820 images were used in this study. We adapted the grading in AREDS dbGaP, which is based on the AREDS severity scale for AMD [38], for reference grading: stage 0 (no AMD), 1 (early AMD), 2 (intermediate AMD), and 3 (advanced AMD, with presence of foveal geographic atrophy (GA) or choroidal neovascularization (CNV)). Categories 0 and 1 are considered non-referable AMD; categories 2 and 3, referable AMD.

### B. Interpretability and Weakly-Supervised Lesion-Level Detection With Iterative Visual Evidence Augmentation

DiaretDB1 [39] was used for the assessment of the interpretability and weakly-supervised detection of DR abnormalities. This dataset consists of 89 CF images with manually-delineated areas performed by four medical experts. Four different types of DR lesions were annotated: hemorrhages, microaneurysms, hard exudates and soft exudates. As proposed in [39], we defined the reference standard as binary masks containing areas labelled with an average confidence level of 75% between experts.

For the assessment of the localization of AMD lesions, we used CF images from the European Genetic Database (EUGENDA), a large multi-center database for clinical and molecular analysis of AMD [40]. AMD severity is defined for each image according to the Cologne Image Reading Center and Laboratory (CIRCL) protocol [40]. We generated a dataset divided in two groups. The first group consists of 52 images with non-advanced AMD stages [41]. Two trained graders manually outlined all visible drusen (without sub-dividing types) in each image, and the binary masks generated during consensus were used as reference standard. In order to assess lesion detection in advanced AMD cases, we created a second group with 12 images with advanced AMD (6 images with advanced dry AMD and 6 images with advanced wet AMD). One professional grader manually delineated in each image all visible AMD-related lesions. To define the reference standard, we generated two binary masks for each image in this group:

drusen (including hard, soft distinct, soft indistinct and optic disk drusen) and advanced-AMD lesions (including CNV, GA and subretinal hemorrhages). In total, 64 images with manually-annotated abnormalities constituted our EUGENDA dataset.

## IV. Experimental Setup

### A. Image-Level Classification

The DR classifier $\mathcal{F}^{DR}_{cnn}$ was trained on the 80% of the Kaggle training set (28,098 images) and validated on the remaining 20% (7,028 images) for 400 epochs. Regarding training configuration, we used the Adam optimizer [42] with a learning rate of 0.0001; data augmentation and class balancing were applied during the training phase to reduce overfitting.

In order to assess the integration of the proposed iterative visual evidence augmentation with different classification network architectures, we performed an additional validation with the Inception-v3 architecture [43] for the classification task of DR grading. As for this alternative DR classifier, $\mathcal{F}^{DR}_{cnn,iv3}$, a dropout layer (p = 0.5) was placed between the final global average pooling layer and the regression node, and it was trained for 100 epochs with the training configuration used previously.

For AMD classification, we applied five-fold cross-validation: the 4,613 patients in the AREDS dataset were randomly divided in five groups, and all the images of each patient were included in the corresponding group. Each fold had an average number of 26,764 images. Three folds were used for training, one for validation and one for testing, with rotation of the folds. In total, five different classifiers were trained for 80 epochs each, using the previously mentioned training configuration. We selected as $\mathcal{F}^{AMD}_{cnn}$ the model which yielded best performance on its corresponding test fold.

### B. Interpretability and Weakly-Supervised Lesion-Level Detection With Iterative Visual Evidence Augmentation

The images in the DiaretDB1 dataset and in the EUGENDA dataset were classified for DR and AMD severity, respectively, with the corresponding image-level classifier. Images whose disease severity prediction was over $th_{pred}$ were considered as referable cases and consequently eligible for interpretability and evaluation of weakly-supervised lesion detection. Similarly to [13], visual evidence of non-referable predictions does not provide meaningful information, since the proposed augmentation aims to unveil iteratively abnormalities while the prediction decreases until non-referability is reached.

The binary masks with annotated lesions were used to assess if the obtained visual evidence highlighted actual abnormalities, and to compare between initial and augmented visual evidence. Free-response ROC (FROC) curves were used for the evaluation of weakly-supervised lesion localization in each dataset and obtained as follows: the points in the interpretability maps with highest confidence values were iteratively located and a circular area of detection with radius $r$ was defined around. If this area overlapped with any annotated lesion in the reference standard, that lesion was considered a true positive detection; otherwise, a false positive detection. The values of the map within the detection area were then masked out, and each lesion in the reference standard detected as true positive was considered only once. For the localization of DR lesions, we defined $r = 7px$ (1.4% image dimensions); for AMD, $r = 10px$ (1.9% image dimensions). From each FROC curve, we extracted the value of sensitivity averaged over all images at a rate of 10 false positives per image on average (SE@10FP). Data bootstrapping [44] was used to assess statistical significance of the obtained metric in each experiment. We computed the 95% confidence intervals (CI), as well as the p-values between initial and augmented SE@10FP values (ratio between bootstrap dataset samples in which the initial performance does not increase after augmentation and the total number of bootstrap dataset samples).

In order to analyze the adaptability of the proposed iterative augmentation to different interpretability methods, we implemented different visual attribution techniques, included in Table I: saliency [18], guided backpropagation [19], integrated gradients [20], Grad-CAM [21], and Guided Grad-CAM [21]. Regarding Grad-CAM, due to the extremely coarse maps generated by this method when the gradient information from the last convolutional layer is used [21], we used the information from a shallower convolutional layer (when using VGG-16: the output of the the third block's last convolutional layer (*Block 3 conv 3*); when using Inception-v3: the output of the second Inception reduction module (*Mixed 8*)).

TABLE II
Quantitative Results on the Kaggle Test Set in the Leaderboard of the Kaggle DR Detection Competition [35]

| # | Method | $\kappa$ |
|---|---|---|
| 1 | Min-Pooling | 0.84957 |
| 2 | o_O | 0.84478 |
| 3 | Reformed Gamblers | 0.83936 |
| 11 | [RU.nl] AI for an Eye | 0.80536 |
| - | $\mathcal{F}^{DR}_{cnn,iv3}$ | 0.80192 |
| 12 | Ryan Munion | 0.79638 |
| 19 | Bingyuan Liu | 0.76797 |
| - | $\mathcal{F}^{DR}_{cnn}$ | 0.76738 |
| 20 | Ilya Kavalerov | 0.76522 |

Metric: $\kappa$, quadratic Cohen's weighted kappa coefficient.

## V. Results

### A. Image-Level Classification

The DR classifier $\mathcal{F}^{DR}_{cnn}$ obtained an AUC of 0.93, with a SE of 0.86 and SP of 0.88, on the Kaggle test set. The model achieved a $\kappa$ of 0.77 for discrimination between DR stages. For the alternative classifier based on the Inception-v3 architecture, $\mathcal{F}^{DR}_{cnn,iv3}$, AUC on the Kaggle test set was 0.93, SE and SP were 0.86 and 0.90, respectively, and $\kappa$ was 0.80.[1] Table II allows to compare the obtained results on the Kaggle test set with those obtained by other entries in the leaderboard of the Kaggle DR detection competition [35].

[1]The ROC analyses of the DR classifiers can be found in Fig. S1 (available in the supplementary files/multimedia tab).

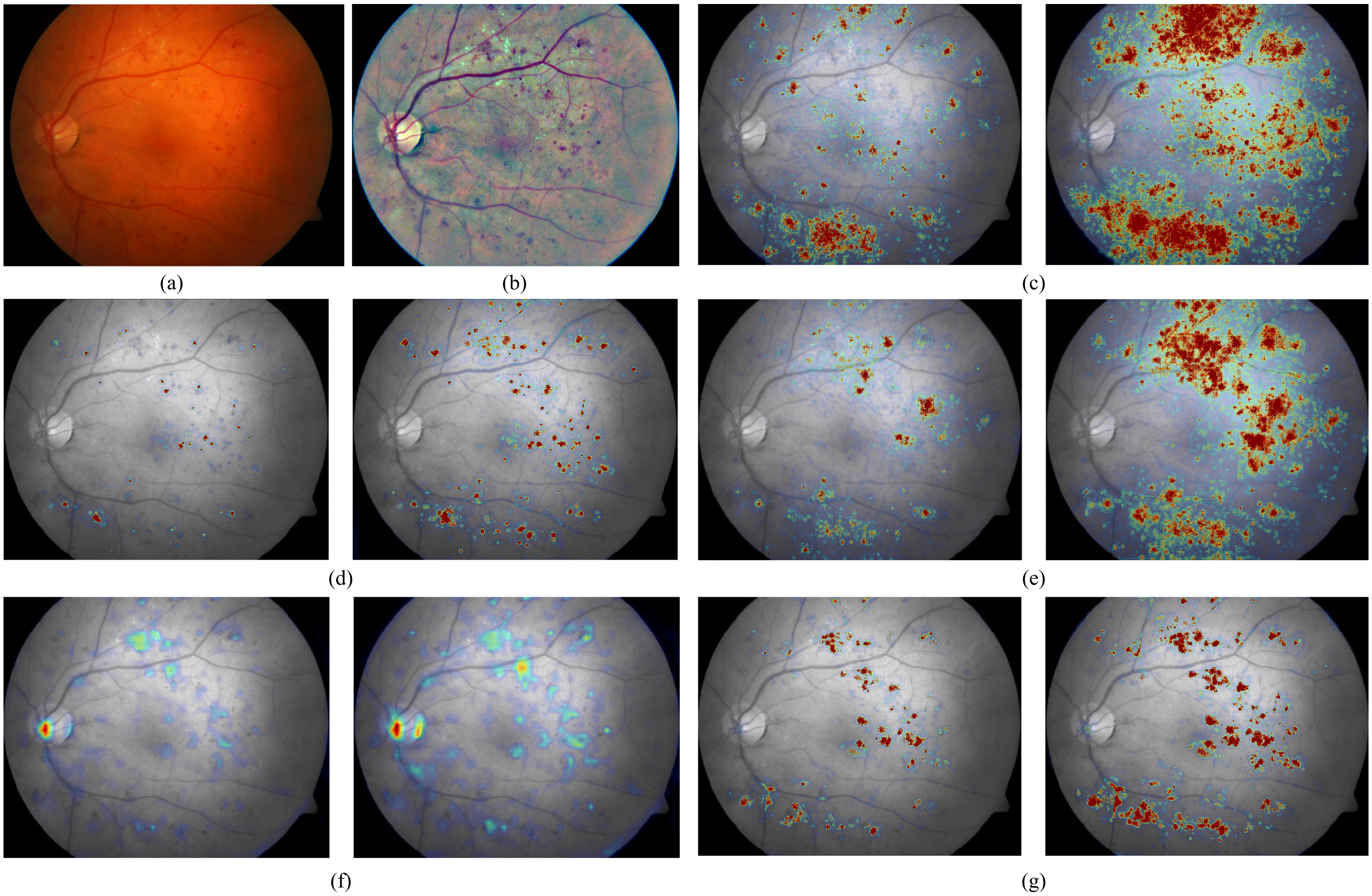

Fig. 2. Example of visual evidence generated with different methods for one image of DiaretDB1, predicted as DR stage 3 with the DR classifier based on VGG-16. For each method: initial visual evidence (left) and augmented visual evidence (right).

TABLE III
QUANTITATIVE RESULTS FOR AMD DETECTION ON THE AREDS SET

| Method | AUC | SE | SP | $\kappa$ | $\kappa_l$ |
|---|---|---|---|---|---|
| DCNN-A WS PP [2] | 0.94 | 0.85 (0.01) | 0.92 (0.01) | - | 0.77 (0.01) |
| Ours (overall performance) | 0.97 | 0.91 (0.01) | 0.92 (0.01) | 0.87 (0.01) | 0.78 (0.01) |

Metrics: AUC, area under the receiver operating characteristic curve; SE, sensitivity; SP, specificity; $\kappa$ and $\kappa_l$, quadratic and linear Cohen's weighted kappa coefficients. Values in parentheses indicate standard deviations. DCCN-A, deep convolutional neural network, algorithm A; WS, with stereo pairs; PP, patient partitioning.

Regarding AMD classification, the overall performance in the AREDS dataset corresponded to an AUC of 0.97, with SE of 0.91 and SP of 0.92 at the optimal operating point; $\kappa$ was 0.87. Table III includes a comparison of the metrics obtained on the whole AREDS set with those obtained in the state of the art [2]. The model with best performance on the corresponding test fold and selected as $\mathcal{F}_{cnn}^{AMD}$ obtained an AUC of 0.97, with SE of 0.92 and SP of 0.93, and a $\kappa$ of 0.88.[2]

[2]The ROC analysis on the whole AREDS set, the ROC analysis of the optimal model, and the performance for each individual model can be found in Fig. S2, Fig. S3 and Table SI (available in the supplementary files/multimedia tab).

## B. Interpretability and Weakly-Supervised Lesion-Level Detection With Iterative Visual Evidence Augmentation

$\mathcal{F}_{cnn}^{DR}$ considered 75 images of the DiaretDB1 to have referable DR. Initial and augmented visual evidence were extracted for these cases. Fig. 2 shows one example from the DiaretDB1 set with the initial and augmented maps for all the implemented visual attribution methods. Table IV includes the quantitative assessment of weakly-supervised localization of four types of DR lesions (hemorrhages, microaneurysms, hard and soft exudates) for the different methods. It contains the SE@10FP values for each type of DR lesion, comparing between initial and augmented visual evidence.[3] Fig. 3 illustrates the FROC curves for the initial and augmented visual evidence per type of lesion generated with guided backpropagation, which is the method that reached the highest average performance, as observed in Table VII.

When $\mathcal{F}_{cnn,iv3}^{DR}$ was used as DR classifier, 67 images in the DiaretDB1 dataset were graded as referable DR. The quantitative results of weakly-supervised detection per DR lesion for the different visual evidence methods can be found in Table V, with and without iterative augmentation.

[3]An additional example from DiaretDB1 for qualitative assessment can be found in Fig. S4 (available in the supplementary files/multimedia tab).

TABLE IV
SE@10FP (95% CI) VALUES FOR WEAKLY-SUPERVISED LESION LOCALIZATION OF DR LESIONS USING VGG-16 ARCHITECTURE

| DR lesion | Visual evidence | Saliency | Guided backpropagation | Integrated gradients | Grad-CAM Block 3 conv 3 | Guided Grad-CAM Block 3 conv 3 |
|---|---|---|---|---|---|---|
| Hemorrhages | Initial | 0.41 (0.31-0.53) | 0.65 (0.54-0.75) | 0.41 (0.32-0.51) | 0.43 (0.28-0.53) | 0.66 (0.51-0.75) |
| | Augmented | 0.50 (0.41-0.61)* | **0.74 (0.67-0.80)*** | 0.58 (0.49-0.68)*** | 0.37 (0.25-0.48) | 0.71 (0.61-0.78)* |
| Microaneurysms | Initial | 0.39 (0.26-0.52) | 0.42 (0.29-0.54) | 0.29 (0.19-0.40) | 0.11 (0.04-0.22) | 0.32 (0.20-0.45) |
| | Augmented | 0.33 (0.22-0.45) | **0.50 (0.37-0.62)*** | 0.25 (0.16-0.35) | 0.11 (0.04-0.21) | 0.37 (0.25-0.50) |
| Hard exudates | Initial | 0.32 (0.22-0.43) | 0.57 (0.42-0.69) | 0.64 (0.49-0.73) | 0.43 (0.31-0.56) | 0.56 (0.44-0.69) |
| | Augmented | 0.38 (0.28-0.49)* | 0.62 (0.50-0.74)* | 0.63 (0.48-0.72) | 0.43 (0.32-0.57) | **0.68 (0.57-0.78)*** |
| Soft exudates | Initial | 0.45 (0.21-0.63) | 0.67 (0.44-0.88) | 0.83 (0.66-0.98) | 0.58 (0.38-0.81) | 0.70 (0.47-0.97) |
| | Augmented | 0.58 (0.36-0.80) | 0.93 (0.71-1.00)*** | **0.98 (0.94-1.00)*** | 0.60 (0.33-0.81) | 0.90 (0.75-1.00)* |

Evaluation performed in cases classified as referable DR in the DiaretDB1 dataset (75/89 images). Shade indicates higher performance after iterative augmentation; bold indicates highest performance per lesion type. P-values: $p \leq 0.05$ (*); $p \leq 0.01$ (**); $p \leq 0.001$ (***).

TABLE V
SE@10FP (95% CI) VALUES FOR WEAKLY-SUPERVISED LESION LOCALIZATION OF DR LESIONS USING INCEPTION-V3 ARCHITECTURE

| DR lesion | Visual evidence | Saliency | Guided backpropagation | Integrated gradients | Grad-CAM Mixed 8 | Guided Grad-CAM Mixed 8 |
|---|---|---|---|---|---|---|
| Hemorrhages | Initial | 0.24 (0.15-0.33) | 0.67 (0.57-0.76) | 0.40 (0.31-0.52) | 0.07 (0.04-0.11) | 0.60 (0.51-0.71) |
| | Augmented | 0.32 (0.23-0.44)*** | **0.76 (0.67-0.83)*** | 0.55 (0.43-0.65)*** | 0.09 (0.04-0.16) | 0.64 (0.51-0.76) |
| Microaneurysms | Initial | 0.23 (0.12-0.36) | 0.55 (0.44-0.68) | 0.24 (0.13-0.37) | 0.00 (0.00-0.00) | 0.50 (0.37-0.63) |
| | Augmented | 0.24 (0.13-0.37) | **0.59 (0.46-0.69)** | 0.22 (0.11-0.37) | 0.03 (0.00-0.10)*** | 0.44 (0.32-0.59) |
| Hard exudates | Initial | 0.17 (0.09-0.26) | 0.58 (0.50-0.73) | **0.71 (0.58-0.85)** | 0.39 (0.27-0.53) | 0.66 (0.54-0.77) |
| | Augmented | 0.20 (0.10-0.29) | 0.62 (0.55-0.78) | 0.70 (0.55-0.84) | 0.31 (0.20-0.42) | 0.65 (0.52-0.81) |
| Soft exudates | Initial | 0.13 (0.00-0.32) | 0.50 (0.23-0.77) | 0.60 (0.38-0.79) | 0.03 (0.00-0.18) | 0.57 (0.35-0.80) |
| | Augmented | 0.23 (0.02-0.45) | 0.73 (0.54-0.93)* | **0.83 (0.71-0.98)** | 0.03 (0.00-0.12) | 0.63 (0.41-0.88) |

Evaluation performed in cases classified as referable DR in the DiaretDB1 dataset (67/89 images). Shade indicates higher performance after iterative augmentation; bold indicates highest performance per lesion type. P-values: $p \leq 0.05$ (*); $p \leq 0.01$ (**); $p \leq 0.001$ (***).

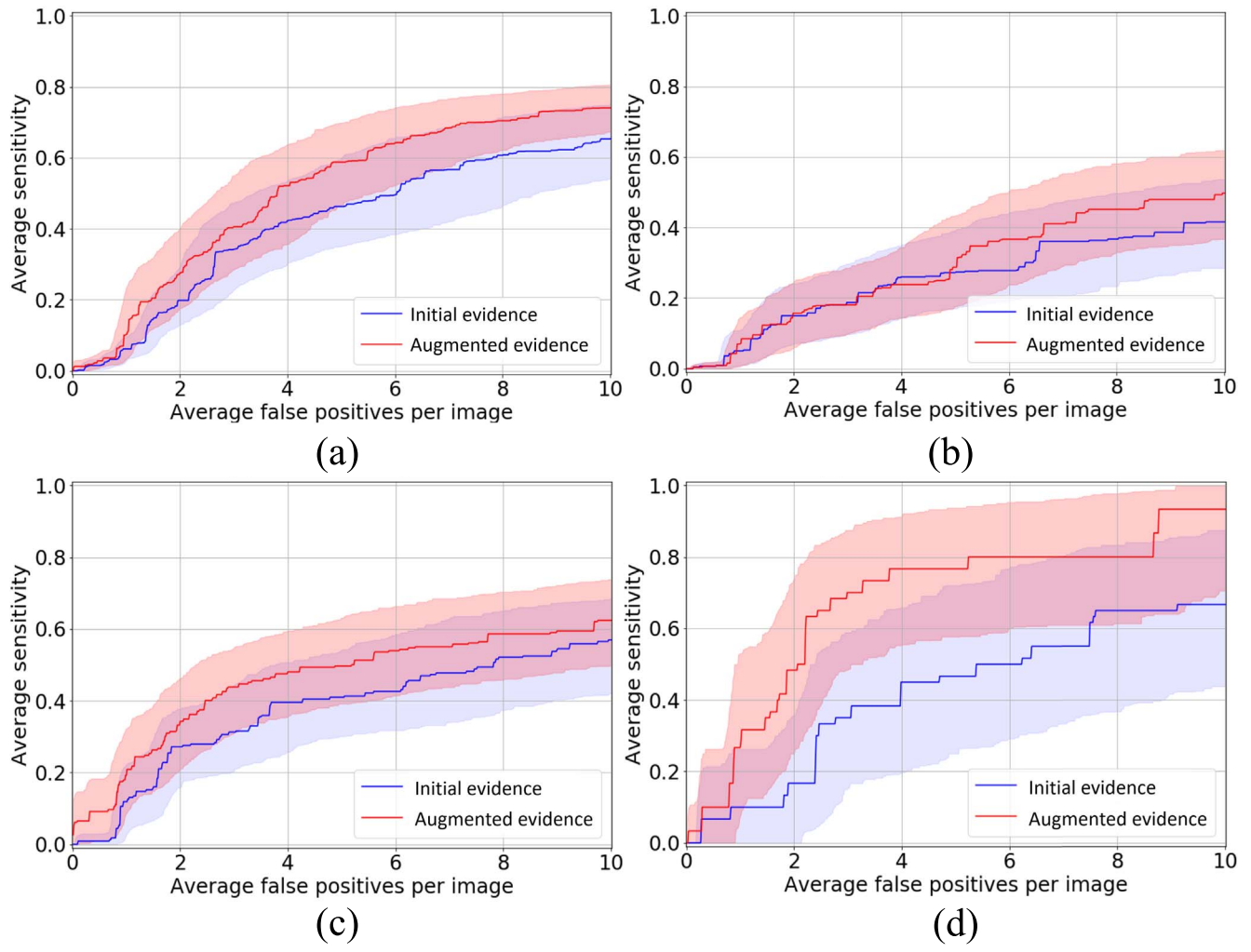

Fig. 3. Lesion localization performance of initial and augmented visual evidence per type of lesion in referable DR predictions from the DiaretDB1 dataset, when using the DR classifier based on VGG-16 and guided backpropagation for visual attribution.

$\mathcal{F}_{cnn}^{AMD}$ graded 40 images in the EUGENDA set as referable AMD. Visual interpretability was extracted for these cases. Fig. 4 includes one example for qualitative evaluation of weakly-supervised AMD lesion localization in this set for all the implemented visual attribution methods, showing the initial and final visual evidence after iterative augmentation. The quantitative assessment of localization of drusen and advanced-AMD lesions can be found in Table VI. In order to analyze the influence of the advanced AMD cases in lesion localization performance, separate quantitative evaluation was carried out on the 52 images with non-advanced AMD stages in the EUGENDA set and results were also included in Table VI.[4]

The global adaptability of the proposed method across classification tasks, network architectures and visual attribution methods can be observed in Table VII. There is a global relative increase of 11.2±2.0% SE@10FP per image in lesion-level localization performance after applying the proposed iterative visual evidence augmentation.

## VI. DISCUSSION

The main highlights of this paper can be summarized as follows:

1) We proposed a novel iterative approach based on the combination of visual attribution and selective inpainting to generate exhaustive interpretability of the predictions made by DL-based classifiers in medical imaging.
2) We introduced for the first time the use of selective inpainting for weakly-supervised lesion localization. We applied selective inpainting on the relevant regions unveiled by visual attribution. Small modifications in the image based on surrounding local information allow to guide the attention of the classifier to new relevant areas in the next iteration.
3) We defined a sensible stopping criterion for the iterative process: the method leads to a gradual decrease in the predicted disease severity until the image is considered as healthy/normal by the classifier.
4) We introduced a method that requires no modifications during or after the training phase of the classifier, neither

[4] An additional example from the EUGENDA set for qualitative assessment can be found in Fig. S5 (available in the supplementary files/multimedia tab).

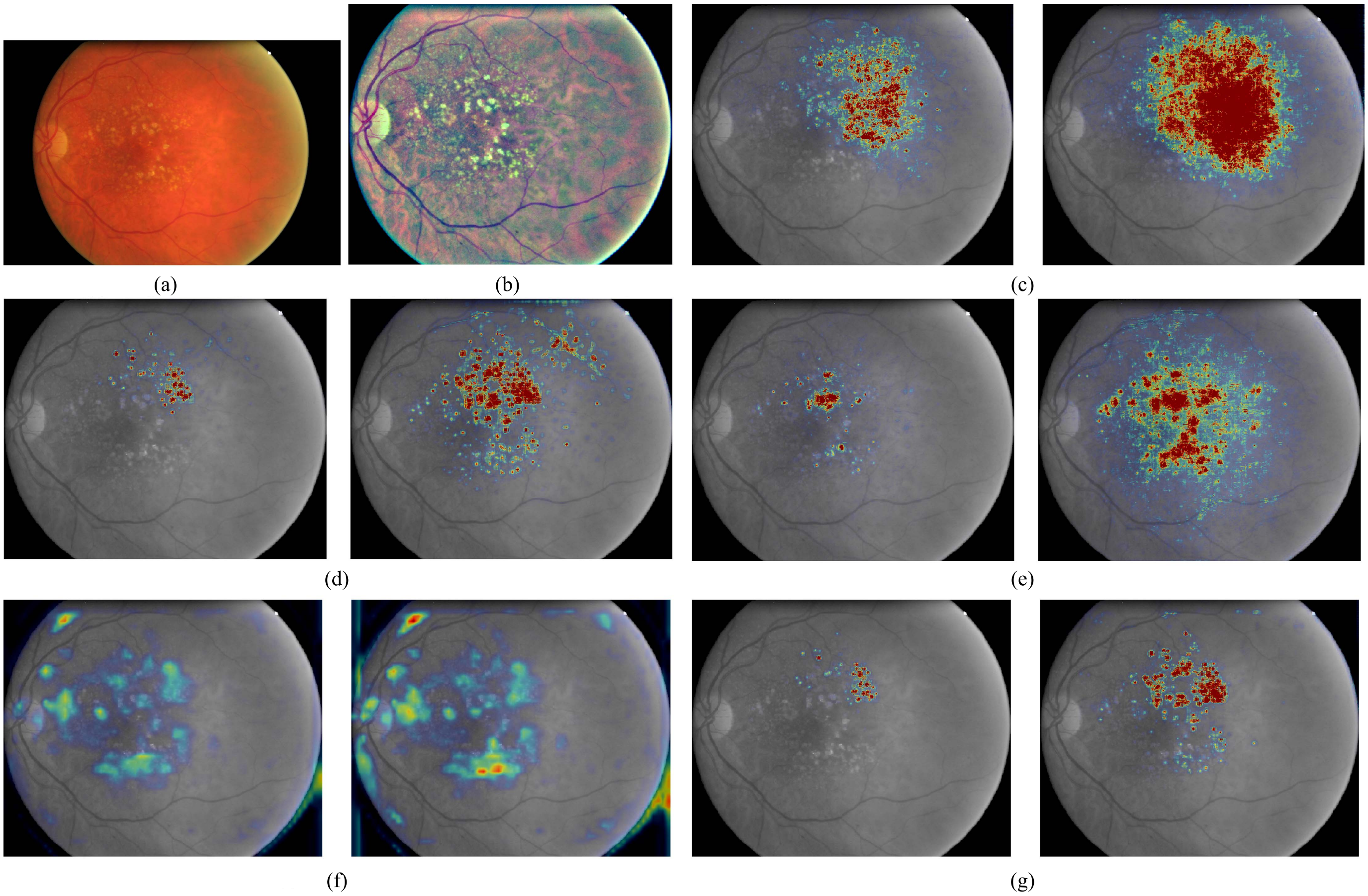

Fig. 4. Example of visual evidence generated with different methods for one image of EUGENDA, predicted as AMD stage 2 (ground-truth label: AMD stage 2). For each method: initial visual evidence (left) and augmented visual evidence (right).

TABLE VI
SE@10FP (95% CI) VALUES FOR WEAKLY-SUPERVISED LESION LOCALIZATION OF AMD LESIONS USING VGG-16 ARCHITECTURE

| DR lesion | Visual evidence | Saliency | Guided backpropagation | Integrated gradients | Grad-CAM Block 3 conv 3 | Guided Grad-CAM Block 3 conv 3 |
|---|---|---|---|---|---|---|
| All EUGENDA (40/64 classified as referable AMD) | | | | | | |
| Drusen | Initial | 0.22 (0.17-0.28) | 0.38 (0.33-0.43) | 0.37 (0.32-0.42) | 0.03 (0.01-0.05) | 0.25 (0.20-0.31) |
| | Augmented | 0.26 (0.21-0.32)*** | 0.44 (0.39-0.50)*** | **0.45 (0.40-0.50)*** ** | 0.02 (0.01-0.03) | 0.26 (0.20-0.33) |
| Advanced lesions | Initial | 0.92 (0.72-1.00) | 0.83 (0.63-1.00) | **0.98 (0.95-1.00)** | 0.27 (0.00-0.50) | 0.88 (0.76-0.97) |
| | Augmented | 0.92 (0.72-1.00) | 0.89 (0.75-1.00) | 0.97 (0.90-1.00) | 0.38 (0.11-0.65) | 0.88 (0.68-1.00) |
| Non-advanced EUGENDA (28/52 classified as referable AMD) | | | | | | |
| Drusen | Initial | 0.22 (0.17-0.30) | 0.42 (0.37-0.48) | 0.40 (0.35-0.46) | 0.02 (0.01-0.05) | 0.27 (0.22-0.34) |
| | Augmented | 0.27 (0.21-0.34)** | **0.51 (0.45-0.57)*** ** | 0.50 (0.44-0.56)*** | 0.01 (0.00-0.02) | 0.28 (0.21-0.37) |

Shade indicates higher performance after iterative augmentation; bold indicates highest performance per lesion type. P-values: $p \leq 0.05$ (*); $p \leq 0.01$ (**); $p \leq 0.001$ (***).

additional integration of postprocessing steps in the interpretability framework.

5) We extensively validated the method for weakly-supervised lesion localization and analyze its adaptability when applied to different classification tasks, visual attribution techniques and network architectures.

Qualitative assessment of the visual evidence generated by the different implemented interpretability methods shows that each DL classifier is able to learn visual features relevant to the classification task at hand during the training process. For those images classified as referable, most visual features correspond to actual abnormalities. The selective inpainting step in the proposed iterative augmentation modifies the abnormal regions highlighted by visual attribution with surrounding local information, guiding therefore the attention of the classifier to new relevant areas in each iteration. This allows to emphasize and refine the delineations of detected abnormalities, as well as to unveil abnormalities that were not highlighted at first but are still related to referable stages and relevant for final diagnosis, independently of anomaly appearance. Consequently, selective inpainting also leads to a gradual decrease in the classifier's disease severity prediction, allowing to stop the iterative augmentation once the classification converges to a healthy prediction. Visual evidence augmentation can be especially observed in severe cases, where the augmented maps differ

TABLE VII
AVERAGE LESION LOCALIZATION PERFORMANCE (SE@10FP (95% CI)) FOR EACH VALIDATED CLASSIFICATION TASK AND NETWORK ARCHITECTURE

| Classification task and architecture | Visual evidence | Saliency | Guided backpropagation | Integrated gradients | Grad-CAM | Guided Grad-CAM |
|---|---|---|---|---|---|---|
| DR, VGG-16 | Initial | 0.39 (0.29-0.48) | 0.58 (0.47-0.66) | 0.54 (0.46-0.61) | 0.39 (0.30-0.48) | 0.56 (0.45-0.67) |
| | Augmented | 0.45 (0.37-0.54)* | **0.70 (0.61-0.75)*** | 0.61 (0.55-0.66)** | 0.38 (0.29-0.47) | 0.67 (0.59-0.73)** |
| DR, Inception-v3 | Initial | 0.19 (0.13-0.27) | 0.58 (0.49-0.69) | 0.49 (0.40-0.58) | 0.12 (0.09-0.18) | 0.58 (0.50-0.67) |
| | Augmented | 0.25 (0.16-0.33) | **0.68 (0.60-0.77)** | 0.58 (0.49-0.68)** | 0.11 (0.08-0.16) | 0.59 (0.50-0.72) |
| AMD, VGG-16 | Initial | 0.57 (0.47-0.63) | 0.61 (0.51-0.70) | 0.68 (0.65-0.70) | 0.15 (0.01-0.27) | 0.57 (0.50-0.62) |
| | Augmented | 0.59 (0.48-0.65)* | 0.67 (0.60-0.72) | **0.71 (0.67-0.74)** | 0.20 (0.06-0.33) | 0.57 (0.46-0.64) |

Shade indicates higher performance across lesion types after iterative augmentation; bold indicates highest performance across lesion types per classification task and architecture. P-values: $p \leq 0.05$ (*); $p \leq 0.01$ (**); $p \leq 0.001$ (***).

more from the initial ones due to a larger number of iterations needed to reach non-referability.

## A. Adaptability of Iterative Visual Evidence Augmentation

As observed in Table VII, the method can be adapted to different visual attribution methods. Nevertheless, it can be observed that iterative augmentation works better when the visual attribution is not coarse, but well localized. Appropriate localization in the initial visual evidence allows to unveil abnormalities of different types, shapes and sizes, such as the ones related to retinal diseases. This can be observed when guided backpropagation is used for visual attribution. Iterative augmentation improves localization performance for AMD lesions (Table VI), as well as for all DR lesions (Table IV, Table V, Fig. 3), where it reaches the highest average performance (Table VII). This corresponds with sharp and localized visual evidence, as observed in Fig. 2 and Fig. 4. Fig. 5 includes additional examples for qualitative assessment of weakly-supervised lesion detection when this method is applied.

On the other hand, as observed in Fig. 2 and Fig. 4, the maps generated using Grad-CAM are hardly detailed, even when a shallower convolutional layer is used for implementation. This was also reported in [15], where CAM were applied with specific fine tuning to improve DR lesions localization. Higher level of coarseness prevents these methods from being a suitable option for interpretability of classification tasks that require precise lesion localization and, in these cases, augmentation does not help, as shown also quantitatively in Tables IV, V and VI. Guided Grad-CAM, due to the combination with guided backpropagation, provides more localized visual evidence and good detection performance especially for most DR lesions, although not better than using guided backpropagation alone, as seen in Table VII.

As for saliency maps, which are more localized than Grad-CAM, augmentation shows visually and quantitatively improvement for detection of most lesions, although final sensitivity values are not high. These maps were used in [11], but adjustment of the training loss and customized, complex postprocessing steps were required to reduce the inherent noise.

Integrated gradients yields better general performance than saliency and Grad-CAM, but maps are more noisy than those obtained with guided backpropagation. Iterative augmentation enhances the localization of AMD lesions, reaching the highest average performance, as seen in Table VII, and certain DR lesions. However, the coarseness and noise of the maps hinders the augmentation's performance for extremely small lesions, such as microaneurysms. Integrated gradients was used in [13], showing support for DR graders, improving confidence and time on task, although no quantitative results of lesion localization were included.

Regarding the adaptability of the proposed method to different architectures, the results in Table V show that weakly-supervised localization of lesions can be generated with different and deeper networks, such as Inception-v3, and improved by means of iterative augmentation.

## B. Quantitative Validation of Weakly-Supervised Lesion-Level Localization

When it comes to quantitative validation for weakly-supervised DR abnormalities detection, only a few of the proposed interpretability frameworks in the literature perform it, such as the one proposed by Quellec et al [11], even though it allows to better understand if the generated visual evidence contains the biomarkers considered by the experts for diagnosis. Additionally, the detection criteria used to generate FROC curves, i.e., interpretation of true positive and false positive detections, usually differs across validation studies. Nevertheless, these curves still allow for certain comparison among methodologies. For our quantitative analysis, we generated FROC curves and extracted the corresponding SE@10FP values as a metric indicative for performance, with additional data boostrapping to analyze the statistical significance of the obtained values.

Table VII shows that the proposed iterative augmentation improves detection across averaged DR lesions for different visual attribution methods and network architectures. Regarding specific lesion detection, Tables IV and V show that augmentation yields improved localization especially for hemorrhages and soft exudates, reaching higher SE@10FP values than those obtained in [11] (hemorrhages: 0.71, soft exudates: 0.90). As for hard exudates, augmentation only improves localization for certain attribution techniques, and a SE@10FP higher than the highest value reached in our paper is achieved in [11] (0.80). Although augmentation also shows improvement in localization of microaneurysms for some techniques, these lesions remain harder to detect, mainly due to its extremely small size. This was also quantitatively shown in [11], although their method achieved higher SE@10FP (0.61). The higher detection performance of [11] for some of the lesions might be due to the use of an ensemble of classifiers optimized for the weakly-supervised detection of each lesion type. However, their validation showed that customized modifications in the classification's training procedure and additional postprocessing were required to improve the visual evidence generated by their framework. This decouples the generated maps from the interpretability of the classifier's prediction. Our method, on the other hand, can be seamlessly integrated in interpretability frameworks, without customized training or postprocessing steps, with varying visual attribution techniques and network architectures.

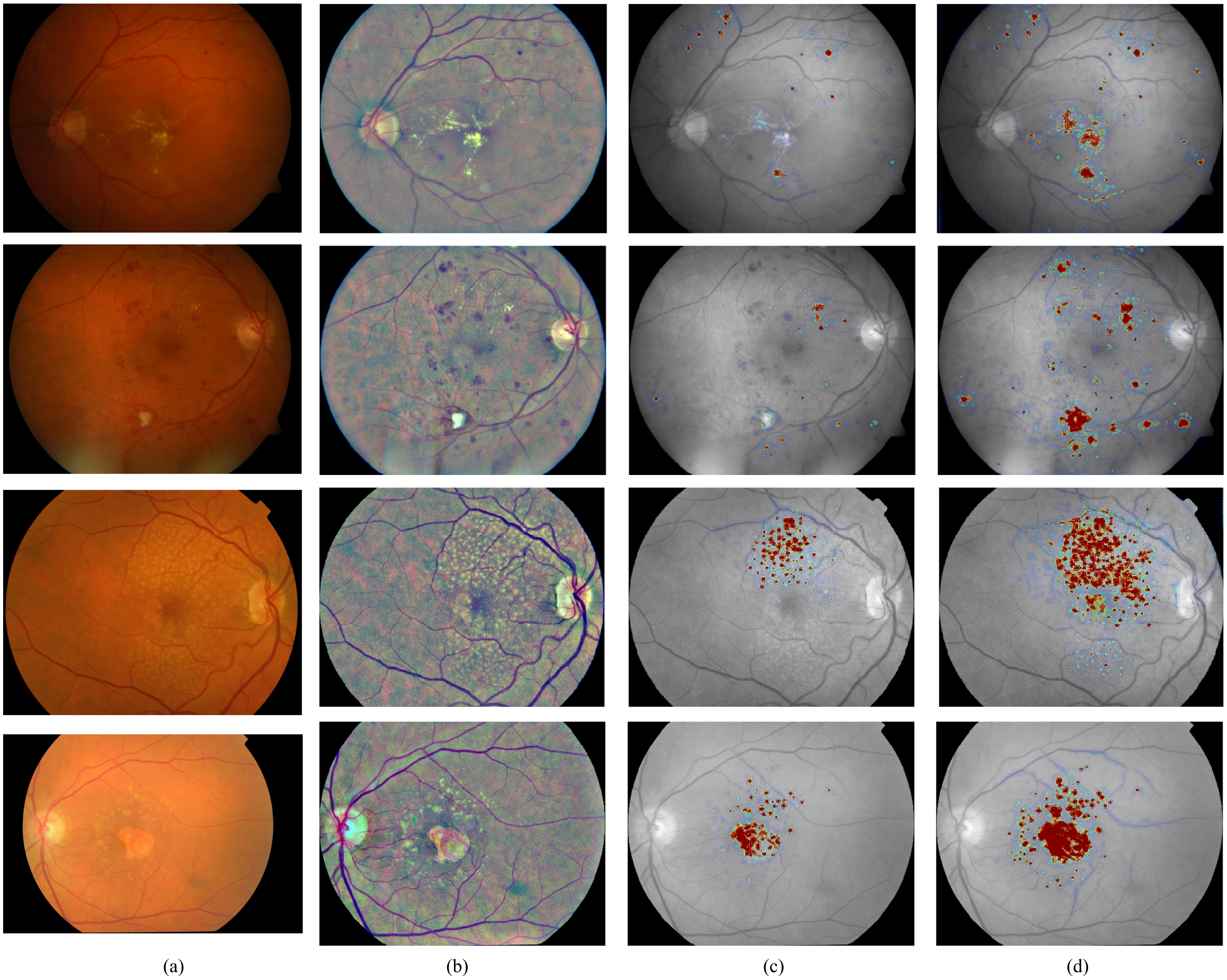

Fig. 5. Examples of visual evidence generated with guided backpropagation for two images in the DiaretDB1 dataset, predicted as DR stage 2 (first row) and 3 (second row), and for two images in the EUGENDA dataset, predicted as AMD stage 2 (third row; ground-truth label: AMD stage 2) and predicted as AMD stage 3 (fourth row; ground-truth label: AMD stage 3).

To the extent of our knowledge, we provide the first quantitative evaluation of weakly-supervised localization of AMD lesions in CF images. As observed in Table VI, advanced-AMD lesions, which should never be missed in grading settings, are initially detected with most interpretability techniques. Augmentation improves drusen detection, although general performance is lower than for DR lesions. This might be related to different aspects. On one hand, AMD grading and annotation of related lesions pose several difficulties to human experts [45], which transfers to the training of DL systems. On the other hand, there is a wide variety of drusen types [46] that are grouped in the presented validation. Table VI illustrates improvement in drusen detection when advanced cases are excluded, i.e., drusen present in advanced AMD stages are harder to unveil, as well as harder for experts to grade [45]. Interpretability of AMD detection will benefit from a validation with further differentiation of drusen types. This would help identify classification burdens and consequent aspects for training optimization.

### C. Limitations and Future Work

Regarding the selective inpainting step in the proposed method, when it comes to small and compact lesions, the surrounding information used to inpaint usually belongs to healthy areas of the image, facilitating the inpainting process and the classification's convergence. On the other hand, the neighbouring areas of large and diffuse abnormalities tend to include both healthy and unhealthy information.

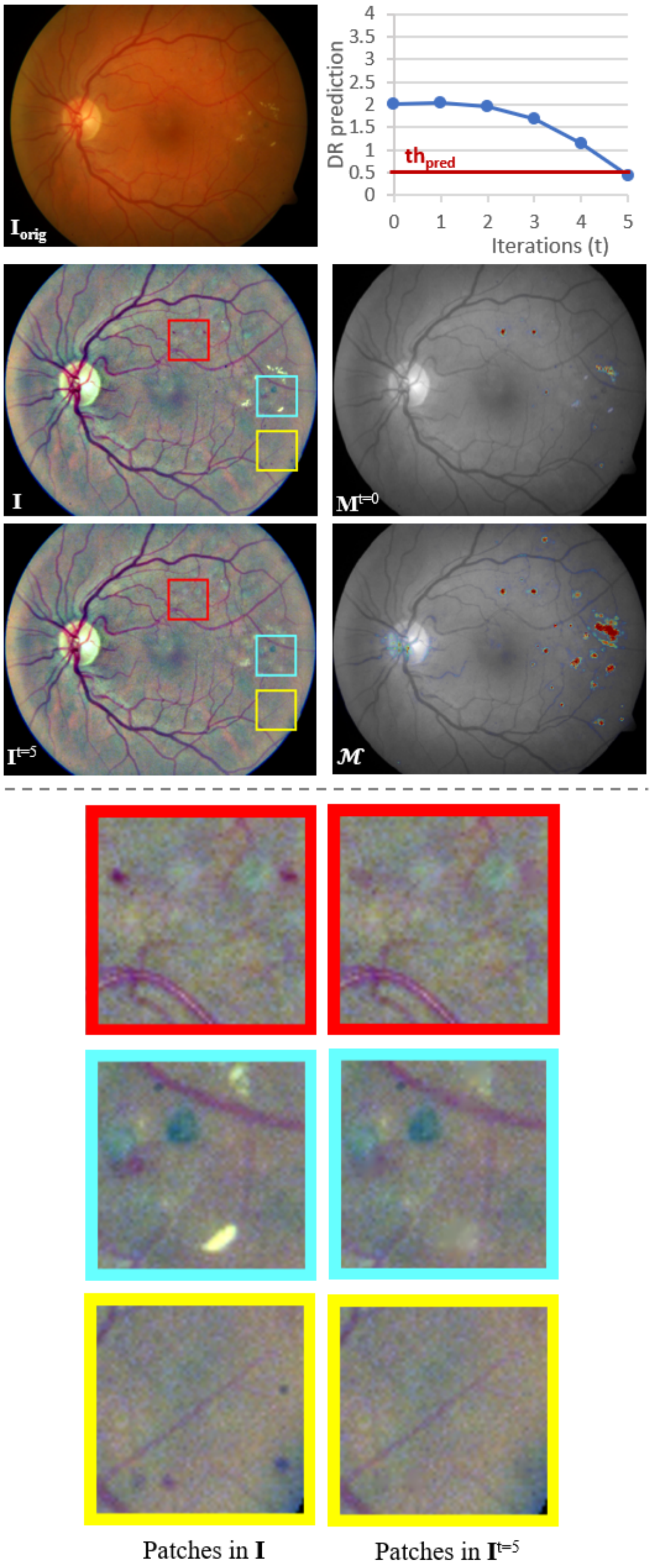

Fig. 6. Detailed comparison between the input image and the resulting inpainted image after applying iterative augmentation to the visual evidence generated for an image in the DiaretDB1 dataset using guided backpropagation as attribution method. $\mathbf{I}_{orig}$, original image; $\mathbf{I}$, input image; $th_{pred}$, prediction threshold; $t$, number of iteration; $\mathbf{M}^t$, explanation map at iteration $t$; $\mathbf{I}^t$, inpainted image at iteration $t$; $\mathcal{M}$, augmented explanation map.

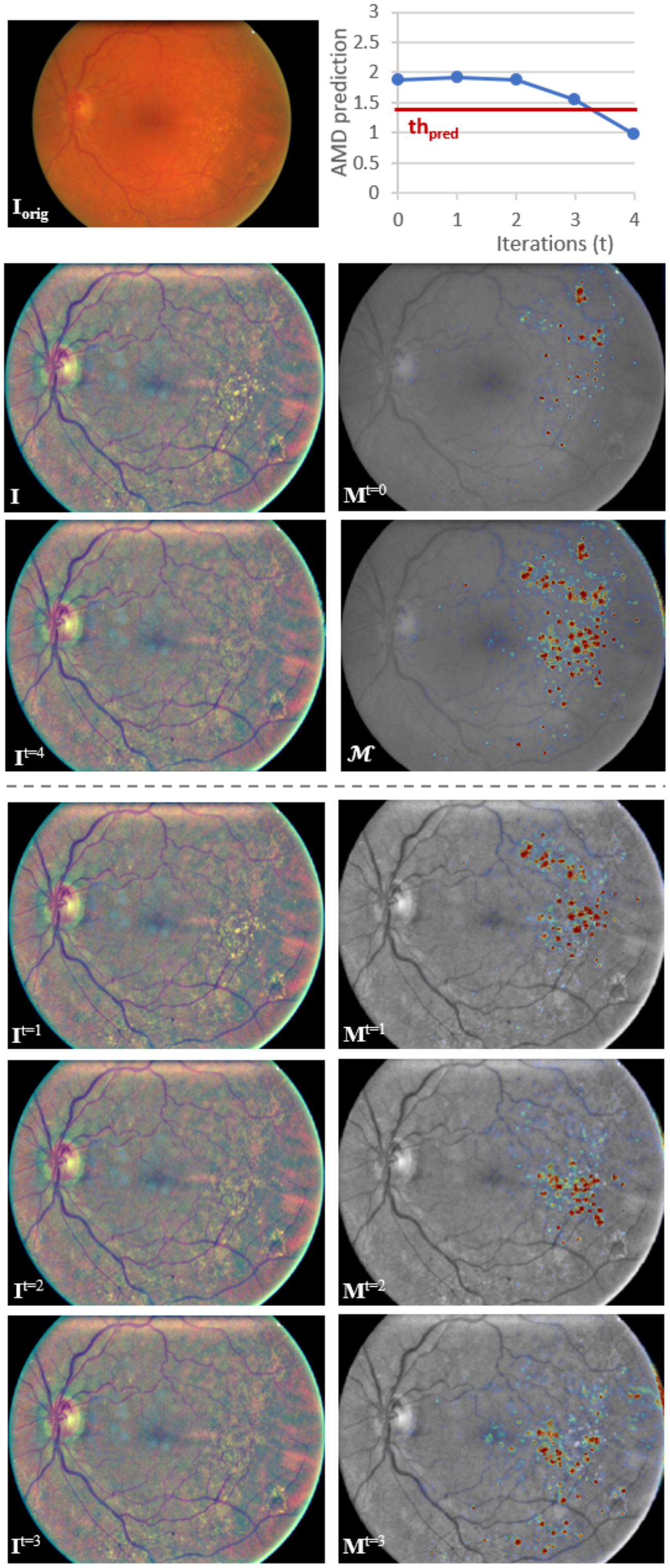

Fig. 7. Visualization of the iterative augmentation process applied to the visual evidence generated for an image in the EUGENDA dataset using guided backpropagation as attribution method. $\mathbf{I}_{orig}$, original image; $\mathbf{I}$, input image; $th_{pred}$, prediction threshold; $t$, number of iteration; $\mathbf{M}^t$, explanation map at iteration $t$; $\mathbf{I}^t$, inpainted image at iteration $t$; $\mathcal{M}$, augmented explanation map.

The classifier's attention is then maintained on a given region for several iterations, which contributes to a refinement in lesion detection, but might also introduce certain oversegmentation. Regarding the classification's convergence, this results in an increased required number of iterations, or lack of convergence, in the case of images with large, advanced lesions.

Another limitation of using selective inpainting is the impossibility to ensure if the modified images are within the original distribution used to train the classifier. Currently, we assume that the training set includes enough variability so as to allow the small local changes in the images introduced by this process to fall within its distribution and the classifier's predictions to be meaningful. As an indicator, we can observe the proportion of images that converge to a healthy prediction after applying the proposed iterative process. Regarding DR detection, 83% of the processed referable images converged to a healthy prediction, averaged across visual attribution methods and network architectures. As for AMD detection, an average of 70% images converged, increasing to 74% when advanced AMD cases were excluded. Fig. 6 and Fig. 7 allow to analyze this aspect qualitatively, by visually inspecting the inpainted images after each iteration and at the end of the process.[5] However, future work would benefit from integrating techniques, such as approaches to leverage the uncertainty in the classifier's decisions [47], to quantitatively determine if inpainted images fall within the original distribution and ensure the classifier's predictions are meaningful.

In this work, we used an unsupervised inpainting technique [28] which yielded satisfactory visual results and fast processing times during iterative augmentation. Future work might include more advanced inpainting techniques, at pixel-level or patch-level, or also trainable with healthy images, such as generative models [48] or context encoders [49].

There are other methods for visual evidence that we have not implemented but that might be interesting to consider for future comparison and integration of iterative augmentation. For instance, layer-wise relevance propagation and its variants [50], [51]. They can be directly applied to a trained classifier to extract interpretability of the predictions and might benefit from iterative augmentation.

Although the proposed method allows to generate an augmented map of visual evidence agnostic to anomaly type and appearance for each prediction, differentiation among detected abnormalities can be useful for a complete explainable diagnosis. In [12], saliency maps were extracted from three different AMD-related classifiers (presence of late-AMD, drusen, and pigmentary abnormalities), yielding one interpretability map per classification task. An ensemble of classifiers for DR grading was used in [11], where one model provided the final DR grade and other models were optimized to provide a map for a given DR lesion type. These solutions allow for separate and optimized interpretability of predictions related to disease grading with respect to a certain lesion type. However, each input image must be processed several times and with multiple maps there is no global and direct interpretation of the actual disease classification. In the future, interpretability of a given classification task will benefit from using the knowledge contained in the corresponding trained network also for differentiation of the lesions included in the visual evidence maps.

The integration of other techniques might improve the usability of the proposed method and help increase trust in the output of the DL classifiers where applied. For example, quantifying and providing information about the uncertainty of the system's decisions [47], as previously mentioned, or exploiting the features learned by the system not only for visual evidence of decisions but also for semantic interpretation [52]. This would allow for better understanding of the features learned by the classifier in the training process and their impact on the final predictions, leading to identify different types of lesions and how they relate to disease severity, as well as new biomarkers significant for disease diagnosis.

## VII. Conclusion

We proposed a deep visualization method for exhaustive visual interpretability of DL classification tasks in medical imaging. The method allows to iteratively increase attention to less discriminative areas that should be considered for final diagnosis, while being adaptable to different classification tasks, network architectures and visual attribution techniques. We showed that visual evidence of the predictions can achieve weakly-supervised lesion-level detection and include the biomarkers considered by the experts for diagnosis. Augmented visual evidence improves the final detection performance, being agnostic to anomaly type and appearance and performing better with sharp and localized initial visual attribution. This makes the proposed method a useful tool for supporting the decisions of medical DL-based classification systems, in order to increase the experts' trust and facilitate their final integration in clinical settings.

[5]Additional examples from the DiaretDB1 and EUGENDA sets for qualitative assessment can be found in Fig. S6-Fig. S9 (available in the supplementary files/multimedia tab).